%% file: acl_latex.tex
\pdfoutput=1

\documentclass[11pt]{article}

\usepackage[preprint]{acl}

\usepackage{times}
\usepackage{latexsym}

\usepackage[T1]{fontenc}

\usepackage[utf8]{inputenc}

\usepackage{microtype}

\usepackage{inconsolata}

\usepackage{graphicx}

\usepackage{amsmath}
\usepackage{multirow}
\usepackage{caption}
\usepackage{subcaption}

\title{ClusterUCB: Efficient Gradient-Based Data Selection for Targeted Fine-Tuning of LLMs}

\author{
 \textbf{Zige Wang\textsuperscript{1}},
 \textbf{Qi Zhu},
 \textbf{Fei Mi},
 \textbf{Minghui Xu},
 \textbf{Ruochun Jin\textsuperscript{2}},
 \textbf{Wenjing Yang\textsuperscript{2*}}
\\
\\
 \textsuperscript{1}School of Computer Science, Peking University, \\
 \textsuperscript{2}College of Computer Science and Technology, National University of Defense Technology
\\
 \small{
   \textbf{Correspondence:} \href{wenjing.yang@nudt.edu.cn}{wenjing.yang@nudt.edu.cn}
 }
}

\begin{document}
\maketitle
\begin{abstract}
Gradient-based data influence approximation has been leveraged to select useful data samples in the supervised fine-tuning of large language models. However, the computation of gradients throughout the fine-tuning process requires too many resources to be feasible in practice. In this paper, we propose an efficient gradient-based data selection framework with clustering and a modified Upper Confidence Bound (UCB) algorithm. Based on the intuition that data samples with similar gradient features will have similar influences, we first perform clustering on the training data pool. Then, we frame the inter-cluster data selection as a constrained computing budget allocation problem and consider it a multi-armed bandit problem. A modified UCB algorithm is leveraged to solve this problem. Specifically, during the iterative sampling process, historical data influence information is recorded to directly estimate the distributions of each cluster, and a cold start is adopted to balance exploration and exploitation. Experimental results on various benchmarks show that our proposed framework, ClusterUCB, can achieve comparable results to the original gradient-based data selection methods while greatly reducing computing consumption.
\end{abstract}

\input{latex/introduction}

\input{latex/related_work}

\input{latex/method}

\input{latex/experiments}

\input{latex/conclusion}


\input{acl_latex.bbl}
\appendix

\input{latex/appendix}

\end{document}

%% file: latex/introduction.tex
\section{Introduction}

Data selection has been a challenging problem in the Supervised Fine-Tuning (SFT) of Large Language Models (LLMs)~\cite{wang2023data, albalak2024survey}. Some researchers propose to use the data influence approximation~\cite{hampel1974influence} to select data samples with the highest influence on target loss optimization during the training process~\cite{charpiat2019input, pruthi2020estimating, xia2024less, wang5206107dynamic}. The data influence at a certain training step is approximated as the inner products or the cosine similarities of the gradients of the training and target validation data samples. Then, the one-step data influence approximations are computed after every short period and aggregated through the training process.

Although proven to be effective, the calculation of data influence approximation consumes many computing resources and can be infeasible in practice when the computing budget is restricted. To reduce resource consumption, one simple way is to compute the gradients of all data samples after longer training periods. Previous work~\cite{wang5206107dynamic} shows that the data influence approximated with gradients will soon lose its indication after multiple training steps. Hence, simply extending the interval of every two times of gradient computation is likely to result in inferior selected data subsets. 

Another way to lower the computation cost is to reduce the number of data samples needed in the calculation of one-step data influence approximation while maintaining the ability to select the data samples with the highest influences. From the derivation of data influence approximation, we come to an intuition that the training data samples with similar gradients tend to have similar influences on the same target loss optimization. Hence, we first perform clustering on all training data samples at the beginning of training according to the similarities of their gradients. In this way, data samples with high influence tend to be concentrated into a few clusters. By picking out the clusters with higher probabilities to contain high-influence data samples, we can avoid the calculation over a large number of low-influence training data samples.

With a constrained computing budget, we frame the data selection among clusters as a computing budget allocation problem, which we call the inter-cluster data selection.
However, a challenge lies in the unknown influence distribution of each cluster. To tackle this challenge, we consider it as a multi-armed bandit problem with each cluster as one arm, and the reward of drawing this arm is the influence of a randomly chosen sample (without replacement) from this cluster. Then, we adapt the most commonly used Upper Confidence Bound (UCB) algorithm to the inter-cluster data selection problem. To better suit our objective, we record the historical rewards and directly estimate the distribution of each arm. A cold start is added to improve the initial estimations. Combining clustering with the modified UCB algorithm, we propose an efficient gradient-based data selection framework \textbf{ClusterUCB} that can be applied to various gradient-based data selection methods. To verify the effectiveness of ClusterUCB, we evaluate it on four widely used benchmarks and two state-of-the-art gradient-based data selection methods~\cite{xia2024less, wang5206107dynamic}. Experimental results demonstrate that ClusterUCB can achieve comparable results with the original gradient-based data selection methods while greatly reducing the computing consumption.

%% file: latex/related_work.tex
\section{Related Work}

\paragraph{SFT Data Selection of LLMs} Selecting suitable data samples for the supervised fine-tuning of large language models has been an ongoing hot topic in the research community~\cite{wang2023data, albalak2024survey}. Extended works are proposed to address different issues of SFT data selection, such as data quality~\cite{zhou2023lima, cao2023instruction, lu2023self}, diversity~\cite{lu2023instag, wan2023explore, ding2023enhancing}, complexity~\cite{he2024complex, zhao2023preliminary}, and so on. While some works try to improve multi-task SFT through proper task composition~\cite{dong2023abilities, Kung2023ActiveIT}, there are also works dedicated to the data selection problem in targeted SFT focusing on a single task~\cite{chen2024skill, xia2024less, wang5206107dynamic}. In these works, gradient-based data influence approximation is used to evaluate the value of individual data samples in the process of targeted SFT. For example, \textit{LESS}~\cite{xia2024less} adapts the classic data influence approximation to Adam optimizer and LoRA~\cite{hu2022lora} training. \textit{Dynamic}~\cite{wang5206107dynamic} points to the decreasing effectiveness of selection during the long-time training process and proposes dynamically updating the selected data coresets. Although effective, these proposed gradient-based data selection methods require high computing consumption in gradient computation of a large number of data samples. Our work addresses this limitation and proposes an efficient gradient-based data selection framework with clustering and a modified UCB algorithm.

\paragraph{Individual data influence} Data influence function is proposed to evaluate the influence of data samples on model training~\cite{hampel1974influence}. Since the evaluation of different combinations of data samples is too costly, some researchers tend to evaluate the influence of individual data samples and treat the sum of these influences as the influence of a data subset. There are two branches in the individual data influence approximation: one is auxiliary model learning and simulation~\cite{ilyas2022datamodels, guu2023simfluence, liu2024training, pmlr-v235-engstrom24a}, and the other is gradient-based training dynamic approximation~\cite{charpiat2019input, pruthi2020estimating, xia2024less, wang5206107dynamic}. In our work, we propose a framework to efficiently apply gradient-based individual data influence approximation in the SFT data selection of LLMs.

%% file: latex/method.tex
\section{Methodology}

\begin{figure*}
    \centering
    \includegraphics[width=1\linewidth]{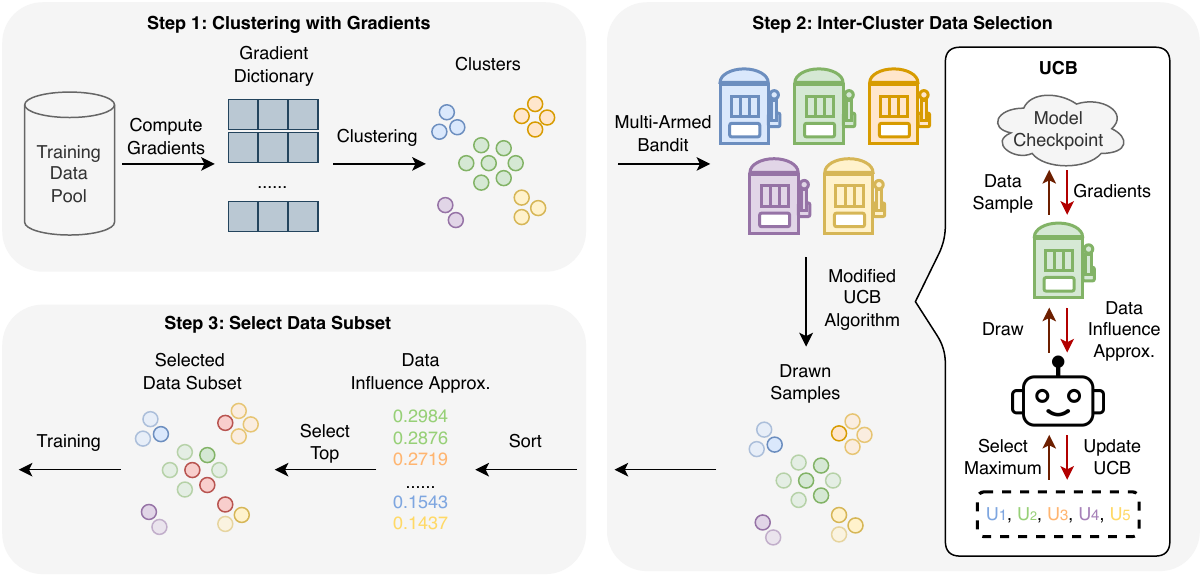}
    \caption{Illustration of ClusterUCB. Step 1: We first compute gradients of all training data samples and perform clustering according to their cosine similarities. Step 2: We frame the inter-cluster data selection as a multi-armed bandit problem, and treat each cluster as one arm. The data influence approximations are considered as the drawing rewards. A modified UCB algorithm is used to draw samples with limited computing budgets. Step 3: From the drawn samples, we select the top portion of data samples with the highest influence as the selected data subset.}
    \label{fig:framework}
\end{figure*}

Our work focuses on the efficient data selection for supervised fine-tuning of large language models on a target task. Given a pretrained model with parameter $\boldsymbol{\theta}$, the training loss $\mathcal{L}(\cdot; \boldsymbol{\theta})$, the training data pool $\mathbf{X}_{tr}=\{ \mathbf{x}_{tr}^1, \mathbf{x}_{tr}^2, ..., \mathbf{x}_{tr}^N\}$, the target task $T$, and the validation data samples representing the target task $\mathbf{X}_{v}=\{ \mathbf{x}_{v}^1, \mathbf{x}_{v}^2, ..., \mathbf{x}_{v}^M\}$, our goal is to select the training data subset with the highest data influence on the training process targeting task $T$ under the constraint of limited computing budget.

\subsection{Preliminary}
Before elaborating on our proposed framework, we first introduce the calculation of gradient-based data influence approximation and the previously proposed data selection methods based on it.

\subsubsection{Gradient-based data influence approximation}
At time step $t$ in the training process, the model parameter is $\boldsymbol{\theta}^t$. Considering the training data sample $\mathbf{x}_{tr}^i$ and validation data sample $\mathbf{x}_{v}^j$, the one-step influence $\mathcal{I}^t$ of $\mathbf{x}_{tr}^i$ is defined as the amount of loss decrease on $\mathbf{x}_{v}^j$ after training on $\mathbf{x}_{tr}^i$ for one step~\cite{pruthi2020estimating}. It can be approximated using the first-order Taylor expansion:
\begin{equation}
\begin{split}
    \mathcal{I}^t(\mathbf{x}_{tr}^i , \mathbf{x}_v^j) & = \mathcal{L}(\mathbf{x}_{v}^j; \boldsymbol{\theta}^{t+1}) - \mathcal{L}(\mathbf{x}_{v}^j; \boldsymbol{\theta}^{t}) \\ & \approx \langle \nabla \mathcal{L}(\mathbf{x}_{v}^j; \boldsymbol{\theta}^{t}), (\boldsymbol{\theta}^{t+1} - \boldsymbol{\theta}^{t}) \rangle,
\end{split}
\end{equation}
where $\nabla \mathcal{L}(\mathbf{x}_{v}^j; \boldsymbol{\theta}^{t})$ is the gradient of $\mathbf{x}_{v}^j$ with respect to $\boldsymbol{\theta}^t$, and $\langle \cdot, \cdot \rangle$ is the inner product.

Since Adam optimizer is the most commonly used optimizer in the SFT of LLMs, the parameter update $\boldsymbol{\theta}^{t+1} - \boldsymbol{\theta}^{t}$ can be represented with adapted gradients $\Gamma$ in Adam. Then, the influence can be approximated as:
\begin{equation} \label{eq:inf-adam}
    \mathcal{I}^t(\mathbf{x}_{tr}^i, \mathbf{x}_v^j) \approx -\eta^t \langle \nabla \mathcal{L}(\mathbf{x}_{v}^j; \boldsymbol{\theta}^{t}), \Gamma(\mathbf{x}_{tr}^i; \boldsymbol{\theta}^{t}) \rangle,
\end{equation}
where $\Gamma(\mathbf{x}_{tr}; \boldsymbol{\theta}^{t}) = -\eta^t\frac{\mathbf{m}^t}{\sqrt{\mathbf{v}^t}+\epsilon}$, and $\mathbf{m}^t$ and $\mathbf{v}^t$ are the moving averages of historical gradients and their element-wise square, respectively.

\subsubsection{Gradient-based data selection}
In previous works~\cite{xia2024less, wang5206107dynamic}, the data influence approximation is adopted to select the most beneficial training data samples for the SFT on the target task. As studied in previous work~\cite{wang5206107dynamic}, to mitigate the selection length bias, the gradients in Equation~\ref{eq:inf-adam} are normalized before calculating the inner product as in:

\begin{equation} \label{eq:norm-inf}
    \mathcal{\widetilde{I}}^t(\mathbf{x}_{tr}^i , \mathbf{x}_v^j) \approx \langle \frac{\nabla \mathcal{L}(\mathbf{x}_{v}^j; \boldsymbol{\theta}^{t})}{\| \nabla \mathcal{L}(\mathbf{x}_{v}^j; \boldsymbol{\theta}^{t}) \|}, \frac{\Gamma(\mathbf{x}_{tr}^i; \boldsymbol{\theta}^{t})}{\| \Gamma(\mathbf{x}_{tr}^i; \boldsymbol{\theta}^{t}) \|} \rangle.
\end{equation}

\citet{wang5206107dynamic} point out that the indication of one-step influence approximation has a declining trend during the training process. Thus, they propose to dynamically recompute the data influence approximation and update the selected data subset after every training period. Instead, LESS~\cite{xia2024less} uses a simulation training with a randomly selected data subset to obtain multiple model checkpoints and performs data selection only once with the aggregated one-step data influence approximations over all checkpoints. 

In each selection, the data influence approximation $\mathcal{\widetilde{I}}(\mathbf{x}_{tr}^i, \mathbf{x}_v^j)$ is aggregated over all validation data samples. Specifically, $\mathcal{\widetilde{I}}(\mathbf{x}_{tr}^i, \mathbf{x}_v^j)$ is first averaged within each subtask, then the maximum among subtasks is chosen as the data influence approximation $\mathcal{\widetilde{I}}(\mathbf{x}_{tr}^i)$ over the validation dataset. After that, the top $p\%$ of training data samples with the highest data influence approximations are selected to form the selected data subset. 

\subsection{Reduce computation consumption with clustering}
\label{subsec:cluster}
To reduce the computation consumption in gradient-based data selection, one simple way is to reduce the number of checkpoints used to calculate the one-step data influence approximation, that is, extending the interval between every two calculations. However, the decreasing long-time selection effectiveness phenomenon illustrated in previous work~\cite{wang5206107dynamic} shows that too long intervals will lead to highly inaccurate influence approximation, which potentially hinders the data selection. Hence, we dedicate our efforts to reducing the number of gradients needed to be computed at each calculation of one-step data influence approximation. 

From Equation~\ref{eq:norm-inf}, the one-step data influence approximation $\mathcal{\widetilde{I}}^t(\mathbf{x}_{tr}^i, \mathbf{x}_v^j)$ is the cosine similarity between the gradients of the training data sample $\mathbf{x}_{tr}^i$ and the validation data sample $\mathbf{x}_{v}^j$. Then, an intuition is that \textbf{the gradients of two training data samples with higher cosine similarity will have similar degrees of cosine similarity with the gradient of the same validation data sample}. Based on this intuition, we perform clustering on the gradients of all training data samples according to their cosine similarities with respect to the pretrained model parameter $\boldsymbol{\theta}^0$. Our experimental observations show that the clusters could remain relatively tight through the model training process, which is discussed in Appendix~\ref{app:cluster}. 

Then, the data samples with the highest influence should be concentrated in a few clusters. Considering each cluster as a distribution over the data influences, we can allocate our computing resources to the clusters according to their probability of containing training data samples with high influences. In this way, we avoid the calculation over a large number of low-influence training data samples, saving a large portion of computing resources.

\subsection{Inter-cluster data selection with UCB algorithm}
\label{subsec:ucb}
After obtaining $k$ clusters $\mathbf{C} = \{C_1, C_2, ..., C_k\}$ for training data samples, our next step is to maximize the overall probability of finding the training data samples with the highest influences by allocating the computing budget among different clusters, which we call it inter-cluster data selection problem with constrained computing budget:
\begin{equation} \label{eq:obj}
\begin{split}
    B^* = & \arg\max \sum_{c=1}^kb_cP_{\mathcal{\widetilde{I}}(\mathbf{x}_{tr})\sim \mathbf{P}_c}(\mathcal{\widetilde{I}}(\mathbf{x}_{tr}) \ge T), \\
    s.t. & \sum_{c=1}^kb_c=\mathcal{B}, \\
    & \forall c\in \{1, ...,k\}, 0 \leq b_c \leq |C_c|,
\end{split}
\end{equation}

where $B=\{b_1, b_2, ..., b_k \}$ is an allocation of the computing budget $\mathcal{B}$,  $\mathcal{\widetilde{I}}(\mathbf{x}_{tr})$ is the data influence approximation of $\mathbf{x}_{tr}$, $\mathbf{P}_c$ is the distribution of data influences contained in cluster $C_c$, and $T$ is the lowest influence of the actual top $p\%$ training data samples with the highest influences.

One challenge exists as the distribution $\mathbf{P}_c$ of each cluster is unknown. Hence, the estimation of $\mathbf{P}_c$ needs to be conducted spontaneously with inter-cluster data selection. This problem setting is very similar to the well-known multi-armed bandit problem~\cite{slivkins2019introduction}. Specifically, each cluster can be considered as one arm with an unknown distribution. Once an arm is drawn in each round, a training data sample will be randomly chosen from the corresponding cluster. Then, its data influence approximation will be calculated as the drawing reward in this round. In this way, the number of drawing rounds is the computing budget consumed by the calculation of data influence approximations. Consequently, optimizing the objective in Equation~\ref{eq:obj} means maximizing the total drawing rewards.

To solve this problem, we adapt the commonly used Upper Confidence Bound (UCB) algorithm~\cite{auer2002finite} to it. The core idea of the UCB algorithm is to estimate an upper confidence bound $U_c$ for each arm that corresponds to cluster $C_c$ in our setting. At each drawing round $d$, the cluster $C^*_d$ with the maximum estimated upper confidence bound $U^*_d$ is chosen to be drawn; then, the reward of drawing $C^*_d$ is acquired and used to update $U^*_d$. By repeating this process, the estimated upper confidence bound of cluster $C_d$ will be closer to the actual expected reward of $C_d$. Hence, the cluster $C^*$ with the highest expected reward will be allocated the most computing budget, while the clusters with lower expected rewards will be allocated less computing budget.

In the classic UCB algorithm~\cite{auer2002finite}, the upper confidence bound is the upper bound of the confidence interval for the estimation of the mean of each arm. Since our objective in Equation~\ref{eq:obj} pays more attention to the probability larger than a certain threshold than the mean of the distribution, we record all historical drawing rewards and use them to estimate the distribution of each cluster. Since the actual $T$ is also unknown, directly estimating $P_{\mathcal{\widetilde{I}}(\mathbf{x}_{tr})\sim \mathbf{P}_c}(\mathcal{\widetilde{I}}(\mathbf{x}_{tr}) \ge T)$ could be challenging. 
Instead, we compare $\hat{T}_c$ for each cluster such that $P^c_{\mathcal{\widetilde{I}}(\mathbf{x}_{tr})\sim \mathbf{P}_c}(\mathcal{\widetilde{I}}(\mathbf{x}_{tr}) \ge \hat{T}_c)$ is approximately equal for all clusters. 
In practice, we use the estimated mean $\hat{\mu}_c$ and standard deviation $\hat{\sigma}_c$ from the historical drawing rewards to compute $\hat{T}_c$ and consider it as the upper confidence bound $U_c$ of cluster $C_c$:
\begin{equation}\label{eq:ucb}
    U_c = \hat{T}_c = \hat{\mu}_c + \beta * \hat{\sigma}_c,
\end{equation}
where $\beta$ is a hyperparameter practically set to 1.

At the initial stage of the UCB algorithm, the insufficient historical drawing rewards might result in the bad estimation of the upper confidence bounds and large regrets in the objective optimization. Hence, we apply a cold start in our UCB algorithm, which allocates a small portion $p_{cs}\%$ of the computing budget among all clusters proportional to the cluster size. After the cold start period, the algorithm starts to choose the cluster with the largest estimated upper confidence bound at each drawing round.

\subsection{Efficient gradient-based data selection framework}
Combining clustering and inter-cluster data selection with the UCB algorithm, we propose our efficient gradient-based data selection framework \textbf{ClusterUCB}, as shown in Figure~\ref{fig:framework}. Specifically, this framework first clusters all training data samples according to the cosine similarities of their gradients computed with respect to the pretrained model. At each time of data selection, the inter-cluster data selection with the UCB algorithm will be applied under a predefined computing budget.

Since there are still regrets that exist in the drawing process, the computing budget is usually set to be larger than the number of training data samples needed in the end. As the final step, we sort the influence approximations calculated in inter-cluster data selection from high to low and output the top number of corresponding training data samples as our final selected data subset.

%% file: latex/experiments.tex
\section{Experiments}

\begin{table*}[t]
    \centering
    \begin{tabular}{lccccccc}
      Methods & Budget & MMLU & TydiQA & GSM8k & HumanEval & Avg. & $\Delta$ \\
      \hline
      Random  & - & 45.3 {\tiny (0.5)} & 48.5 {\tiny (0.6)} & 19.7 {\tiny (0.1)} & 16.5 {\tiny (0.8)} & 32.5 & - \\
      \hline 
      LESS   & 100\% & 47.0 {\tiny (0.5)} & 53.2 {\tiny (1.1)} & 27.3 {\tiny (0.6)} & 17.7 {\tiny (0.9)} & 36.3 & - \\
      LESS-Rerank & 20\% & 45.9 {\tiny (0.2)} & 52.2 {\tiny (0.3)} & 23.8 {\tiny (0.1)} & 16.3 {\tiny (0.3)} & 34.6 & $\downarrow$ 1.7 \\
      LESS-ClusterUCB & 20\% & \textbf{47.8 {\tiny (0.8)}} & \textbf{53.7 {\tiny (1.3)}} & \textbf{28.0 {\tiny (1.0)}} & \textbf{17.6 {\tiny (0.7)}} & \textbf{36.8} & \textbf{$\uparrow$ 0.5} \\
      \hline
      Dynamic & 100\% & 47.8 {\tiny (0.3)} & 57.6 {\tiny (0.8)} & 27.5 {\tiny (1.2)} & 19.2 {\tiny (0.4)} & 38.0 & - \\
      Dynamic-Rerank & 20\% & 46.7 {\tiny (0.4)} & 54.9 {\tiny (0.9)} & 26.7 {\tiny (0.6)} & \textbf{17.9 {\tiny (0.7)}} & 36.6 & $\downarrow$ 1.4 \\
      Dynamic-ClusterUCB & 20\% & \textbf{47.9 {\tiny (0.6)}} & \textbf{57.4 {\tiny (0.6)}} & \textbf{27.4 {\tiny (0.9)}} & 17.7 {\tiny (0.4)} & \textbf{37.6} & \textbf{$\downarrow$ 0.4} \\
    \end{tabular}
    \caption{The results of ClusterUCB and baselines on four commonly-used benchmarks. All experiments are repeated with three random seeds. $\Delta$ denotes the difference of average performance between budget-constrained methods and their full-budget counterparts. \textbf{Bold} means the best results achieved by budget-constrained methods.}
    \vspace{-3mm}
    \label{tab:main}
\end{table*}

\subsection{Experimental setup}
\paragraph{Baselines} \textbf{Random} is to train the model with a randomly selected data subset. \textbf{LESS}~\cite{xia2024less} uses simulation training with randomly selected data to acquire multiple model checkpoints, aggregates the one-step data influence approximations of all training data samples with respect to these checkpoints, and chooses the top $p\%$ training data samples with the highest aggregated influence approximations as final selected data subset. \textbf{Dynamic}~\cite{wang5206107dynamic} directly uses one-step data influence approximations to select top $p\%$ training data samples, but the selection process will be repeated periodically, leading to dynamically updated data subsets through the model training process. To show the effectiveness of our modified UCB algorithm with the same computing budgets, we also implement two vanilla baselines \textbf{LESS-Rerank} and \textbf{Dynamic-Rerank}, which randomly choose $\mathcal{B}$ training data samples to compute their influence approximations and select the top $p\%$ high-influence data samples among them.

\paragraph{Implementation details} Following \textit{LESS} and \textit{Dynamic}, We use LLaMA-2-7B~\cite{touvron2023llama2} as our pretrained model and set the selection ratio $p\%$ to 5\%. The pretrained model is trained for four epochs with an AdamW optimizer. The learning rate is 2e-5 with linear decay. For \textit{Random}, 5\% of the training data samples are randomly selected to train the model. For \textit{LESS} and \textit{Dynamic}, the implementation is kept the same as described in their original paper: in \textit{LESS}, 5\% randomly selected training data samples are used for simulation training to obtain four checkpoints after each epoch; in \textit{Dynamic}, one-step data selection is performed at the beginning of each epoch, and 20 warmup steps are performed for the first one-step data selection. For ClusterUCB, we also perform 20 warmup steps and use the resulting checkpoint to compute the gradients of all training data samples. Then, K-means~\cite{hartigan1979algorithm} is adopted for clustering. We combine ClusterUCB with \textit{LESS} and \textit{Dynamic} to form two variants, \textbf{LESS-ClusterUCB} and \textbf{Dynamic-ClusterUCB}, respectively. In \textit{LESS-ClusterUCB}, the reward of each draw is the aggregated one-step data influence approximations as in \textit{LESS}. In \textit{Dynamic-ClusterUCB}, since the gradients used for clustering and the first one-step data selection are the same and complete, we keep the first training epoch the same as that in \textit{Dynamic}, then apply our proposed inter-cluster data selection in the following three one-step data selections. In our main experiments, for \textit{LESS-ClusterUCB}, \textit{Dynamic-ClusterUCB}, \textit{LESS-Rerank}, and \textit{Dynamic-Rerank}, the number of clusters $k$ is 150, the cold start ratio $p_{cs}\%$ is 5\%, and the computing budget $\mathcal{B}$ is 20\% of the total number of training data samples. All gradients are computed using LoRA~\cite{hu2022lora} and projected to 8192-dimensional vectors using Random Projection~\cite{park2023trak}.

\paragraph{Datasets} 
The training data pool is the mix of 8 commonly-used datasets: \textbf{Flan-v2}~\cite{longpre2023flan} is a large SFT dataset converted from various NLP datasets; \textbf{CoT}~\cite{longpre2023flan} is a subset of Flan-v2 with chain-of-though; \textbf{Dolly}~\cite{DatabricksBlog2023DollyV2} is a high-quality instruction-following dataset generated by humans; \textbf{Open Assistant v1}~\cite{kopf2023openassistant} is a multi-round chatting datasets generated by human and open-sourced LLMs; \textbf{GPT4-Alpaca} ~\cite{peng2023instruction} contains instructions in Alpaca dataset and answers regenerated by GPT-4; \textbf{ShareGPT} ~\cite{vicuna2023} is a conversation datasets with mixed-quality; \textbf{GSM8k train}~\cite{cobbe2021training} is a primary school-level math word dataset; \textbf{Code-Alpaca}~\cite{codealpaca} is a dataset designed for the development of model's coding ability.

\paragraph{Evaluation benchmarks and validation data samples} We adopt four commonly used benchmarks covering the general, multilingual, mathematical, and coding abilities of LLMs. \textbf{MMLU}~\cite{hendryckstest2021} is a knowledge-based multi-choice QA benchmark including 57 subjects; \textbf{TydiQA}~\cite{clark2020tydi} is a multi-language QA benchmark including nine languages; \textbf{GSM8k}~\cite{cobbe2021training} is a math reasoning benchmark evaluating models' mathematical reasoning ability; \textbf{HumanEval}~\cite{chen2021codex} is a Python coding benchmark evaluating models' code generation ability. The selection and aggregation of validation data samples follows \textit{Dynamic} for all methods: the few-shot samples of MMLU and TydiQA are directly used as validation data samples; 50 and 10 test data samples are randomly selected from GSM8k and HumanEval as validation data samples.

\subsection{Main results}

The performances of ClusterUCB and baselines on four benchmarks are shown in Table~\ref{tab:main}. All gradient-based methods outperform \textit{Random} to a large margin on the average performance of four benchmarks, showing that gradient-based methods are effective in selecting suitable data subsets for the targeted fine-tuning of LLMs.

With the computing budget set to 20\%, both \textit{LESS-ClusterUCB} and \textit{Dynamic-ClusterUCB} match their full-budget counterparts \textit{LESS} and \textit{Dynamic} on most benchmarks. These results show that ClusterUCB can reduce the computational consumption of gradient-based data selection methods while maintaining their performance. Although the performance of \textit{Dynamic-ClusterUCB} on HumanEval drops compared to \textit{Dynamic}, in Section~\ref{subsec:budget}, we find that it could achieve better results when the computing budget is reduced to 10\%.

Using the same computing budget, both \textit{LESS-ClusterUCB} and \textit{Dynamic-ClusterUCB} outperform \textit{LESS-Rerank} and \textit{Dynamic-Rerank} on almost all benchmarks, further indicating the effectiveness of the selection strategy used in ClusterUCB.

\subsection{Influence of computing budgets}
\label{subsec:budget}

\begin{table}[t]
    \centering
    \begin{tabular}{lccccc}
      Bgt & MMLU & TydiQA & GSM8k & HE \\
      \hline
      10\%  & 46.3 {\tiny (1.1)} & 55.4 {\tiny (0.5)} & 28.8 {\tiny (0.5)} & 18.8 {\tiny (0.6)} \\
      20\%   & 47.9 {\tiny (0.6)} & 57.4 {\tiny (0.6)} & 27.4 {\tiny (0.9)} & 17.7 {\tiny (0.4)} \\
      30\% & 47.3 {\tiny (0.2)} & 57.5 {\tiny (0.5)} & 27.2 {\tiny (1.4)} & 18.3{\tiny (0.3)} \\
    \end{tabular}
    \caption{The results of \textit{Dynamic-ClusterUCB} with different computing budgets. Bgt and HE are the abbreviations for Budget and HumanEval, respectively.}
    \vspace{-3mm}
    \label{tab:budget}
\end{table}

To illustrate the influence of computing budgets, we conduct experiments on \textit{Dynamic-ClusterUCB} with computing budgets $\mathcal{B}$=10\%, 20\%, and 30\%. The cold start ratio $p_{cs}\%$ is 5\%, and the number of clusters is 150, as in the main experiments. The results in Table~\ref{tab:budget} show that different benchmarks have different change patterns along with the computing budget. On MMLU and TydiQA, the performance of \textit{Dynamic-ClusterUCB} is worse when $\mathcal{B}$ is only 10\%. $\mathcal{B}$ = 20\% is enough since increasing $\mathcal{B}$ from 20\% to 30\% leads to trivial performance improvements. On the contrary, on GSM8k and HumanEval, the smaller computing budgets tend to result in higher accuracies.
The reason might be that the training data samples useful for the improvement of mathematical and coding abilities are spread across only a few clusters. Thus, a small computing budget is sufficient to find the data samples with high influence once our UCB algorithm finds the correct arms. The degradation of performance with the increase of computing budgets on GSM8k and HumanEval might lie in the randomness of data selection and training.

\subsection{Results on a different model}
\label{subsec:qwen}

\begin{table}[t]
    \centering
    \begin{tabular}{lccc}
      Methods & TydiQA & HumanEval \\
      \hline
      Random  & 64.5 {\tiny (0.2)} & 38.4 {\tiny (1.2)} \\
      \hline 
      LESS   & 66.8 {\tiny (0.5)} & 36.9 {\tiny (1.5)} \\
      LESS-ClusterUCB & 66.7 {\tiny (0.5)} & 38.1 {\tiny (0.9)}  \\
      \hline
      Dynamic & 67.3 {\tiny (0.6)} & 40.0 {\tiny (1.5)} \\
      Dynamic-ClusterUCB & 67.8 {\tiny (0.1)} & 40.9 {\tiny (1.2)} \\
    \end{tabular}
    \caption{The results of ClusterUCB and baselines using Qwen2.5-3B as the pretrained model.}
    \vspace{-3mm}
    \label{tab:qwen}
\end{table}

To further evaluate ClusterUCB on a different model architecture and scale, we conduct experiments using Qwen2.5-3B~\cite{qwen2.5} as the pretrained model on TydiQA and HumanEval benchmarks. The implementation details are kept the same as in our main experiments, except that all models are trained for three epochs. 

The results are shown in Table~\ref{tab:qwen}. 
Consistent with the results using LLaMA-2-7B as the pretrained model, both \textit{LESS-ClusterUCB} and \textit{Dynamic-ClusterUCB} match their full-budget counterparts \textit{LESS} and \textit{Dynamic} on these two benchmarks, showing the effectiveness of ClusterUCB on different model architectures and scales.

\subsection{Hyperparameter analysis}
We analyze the impacts of two key hyperparameters in ClusterUCB: the number of clusters $k$ and the cold start ratio $p_{cs}\%$. We adopt two metrics to evaluate the goodness of the selected hyperparameters. One is the sample-level recall rate $R_{s}$, and the other one is the influence-level recall rate $R_{inf}$, as shown in Equation~\ref{eq:recall} and ~\ref{eq:score}, respectively, where $D$ is the final data subset selected by ClusterUCB and $D_{gt}$ is the actual top portion of training data samples with the highest influence. 
\begin{equation} \label{eq:recall}
    R_{s}=\frac{|D\bigcap D_{gt}|}{|D_{gt}|}
\end{equation}
\begin{equation} \label{eq:score}
    R_{inf} = \frac{\sum_{\mathbf{x}_{tr}^i\in D}\mathcal{\widetilde{I}}(\mathbf{x}_{tr}^i)}{\sum_{\mathbf{x}_{tr}^q\in D_{gt}}\mathcal{\widetilde{I}}(\mathbf{x}_{tr}^q)}.
\end{equation}

We compute $R_{s}$ and $R_{inf}$ on the model checkpoint obtained after warmup training, and the computing budget $\mathcal{B}$ is fixed to be 20\% in this section. 

\subsubsection{Cold start ratio}
\label{subsubsec:csp}

\begin{figure}
    \centering
    \includegraphics[width=\linewidth]{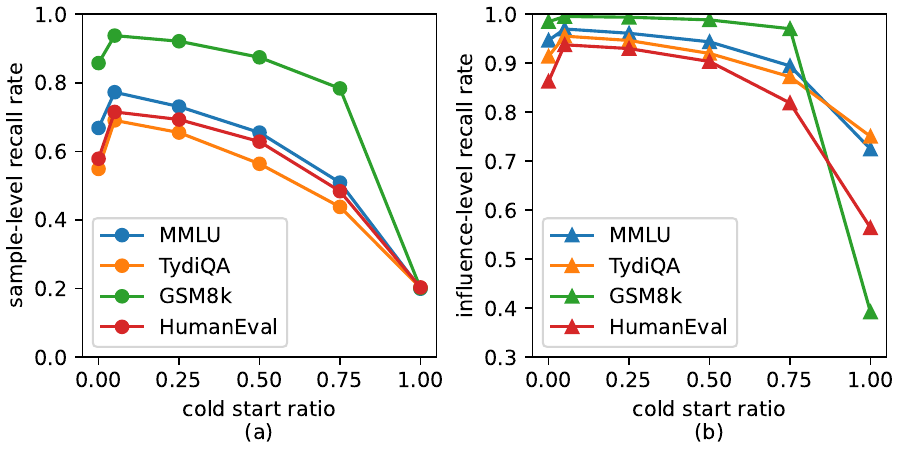}
    \caption{The sample-level and influence-level recall rates with different cold start ratios.}
    \label{fig:csp}
\end{figure}

\begin{figure}
    \centering
    \includegraphics[width=\linewidth]{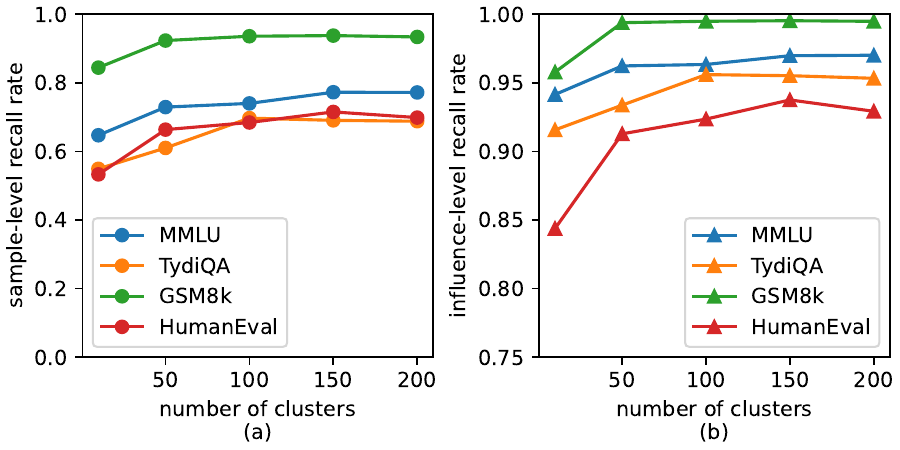}
    \caption{The sample-level and influence-level recall rates with different numbers of clusters.}
    \label{fig:k}
\end{figure}

\begin{table*}
    \centering
    \begin{tabular}{lcccccccccc}
        \hline
          & \multicolumn{2}{c}{Random-Draw} & \multicolumn{2}{c}{UCB1} & \multicolumn{2}{c}{UBC-TN} & \multicolumn{2}{c}{UCB-TH} & \multicolumn{2}{c}{UCB-Beta} \\
        Tasks & $R_{s}$ & $R_{inf}$ & $R_{s}$ & $R_{inf}$ & $R_{s}$ & $R_{inf}$ & $R_{s}$ & $R_{inf}$ & $R_{s}$ & $R_{inf}$ \\
        \hline
        MMLU   & 20.14 & 72.72 & 26.24 & 76.76 & 73.96 & 96.40 & 77.00 & 96.96 & \textbf{77.24} & \textbf{96.97} \\
        TydiQA & 19.38 & 74.74 & 23.12 & 77.21 & 59.08 & 92.96 & 68.96 & \textbf{95.56} & \textbf{69.03} & 95.52 \\
        GSM8k  & 25.84 & 45.87 & 59.31 & 90.05 & 90.64 & 99.13 & 93.12 & 99.47 & \textbf{93.75} & \textbf{99.52} \\
        HumanEval  & 22.63 & 58.68 & 24.72 & 61.01 & 55.96 & 86.95 & \textbf{71.61} & \textbf{93.83} & 71.50 & 93.75 \\
        \hline
    \end{tabular}
    \caption{The sample-level and influence-level recall rates of different upper confidence bound evaluation metrics. \textbf{Bold} indicates the best results for each task.}
    \vspace{-3mm}
    \label{tab:metric}
\end{table*}

Set $k$ = 150,  we evaluate $p_{cs}\%$ = \{0\%, 5\%, 25\%, 50\%, 75\%, 100\%\}, as shown in Figure~\ref{fig:csp}. 

\paragraph{Sufficient historical reward information is necessary.} When $p_{cs}\%$ = 0\%, both $R_{s}$ and $R_{inf}$ are low for all tasks. But with $p_{cs}\%$ increases to 5\%, $R_{s}$ and $R_{inf}$ improve obviously. It shows that sufficient historical reward information with a cold start is necessary in the initial stage of inter-cluster data selection.

\paragraph{Our UCB algorithm is effective in inter-cluster data selection.} When $p_{cs}\%$ increases to 100\%, the inter-cluster data selection degrades to simply allocating computing budgets proportional to the cluster size, and $R_{s}$ and $R_{inf}$ become extremely low. It illustrates that our UCB algorithm is effective in inter-cluster data selection. 

\paragraph{The trade-off between exploration and exploitation.} With the increase of $p_{cs}\%$ from 5\%, $R_{s}$ and $R_{inf}$ gradually decrease, showing that only a small portion of the computing budget should be used for the cold start. The ratio of cold start also controls the trade-off between exploration and exploitation in the UCB algorithm. These results indicate that exploration is necessary and important in our UCB algorithm, but too much exploration could hinder the algorithm's performance. We refer to Appendix~\ref{app:dist} for further discussion about the distribution of data selection among clusters.

\subsubsection{Number of clusters}

Set $p_{cs}\%$ = 5\%, we evaluate $k$ = \{10, 50, 100, 150, 200\}. The results are shown in Figure~\ref{fig:k}. When the number of clusters is as small as 10, both $R_{s}$ and $R_{inf}$ are the worst for all tasks, indicating that too small number of clusters might not be able to fully separate training data samples into groups with similar gradients. Increasing the number of clusters from 10 to 50, $R_{s}$ and $R_{inf}$ show obvious improvement. Further increasing the number of clusters, the improvements become less observable, and $R_{s}$ and $R_{inf}$ tend to be stable. It also indicates that ClusterUCB is not sensitive to the number of clusters, as long as it is not too small.

\subsection{Comparison of different upper confidence bound evaluation metrics}

In Section~\ref{subsec:ucb}, we evaluate the upper confidence bound $U_c$ as the estimated influence threshold $\hat{T}_c$ that corresponds to the same probability. An alternative is to directly estimate $P_{\mathcal{\widetilde{I}}(\mathbf{x}_{tr})\sim \mathbf{P}_c}(\mathcal{\widetilde{I}}(\mathbf{x}_{tr}) \ge T)$. One estimation is the ratio of drawing with rewards larger than $T$, which we call \textbf{UCB-TH}. We could also consider each cluster distribution $\mathbf{P}_c$ as a Gaussian distribution with the mean and standard deviation estimated from the historical reward values $\widetilde{\mathbf{P}}_c \sim \mathcal{N}(\hat{\mu}_c, \hat{\sigma}_c)$, and compute $P_{\mathcal{\widetilde{I}}(\mathbf{x}_{tr})\sim \widetilde{\mathbf{P}}_c}(\mathcal{\widetilde{I}}(\mathbf{x}_{tr}) \ge T)$. We call this estimation \textbf{UCB-TN}. Since $T$ is unknown, we estimate it as the lowest influence in the top $p/\mathcal{B}$ portion of all historical reward values of all clusters in the current round.

Keeping all hyperparameters and computing budget the same as in the main experiments, we compare \textit{UCB-TH} and \textit{UCB-TN} with the evaluation metrics used in our main experiments (\textbf{UCB-Beta}). We also compare them with two baselines: \textbf{Random-Draw} that randomly chooses an arm to draw at each round, and a classic UCB algorithm \textbf{UCB1}~\cite{auer2002finite}. 

The results in Table~\ref{tab:metric} show that \textit{UCB-Beta} and \textit{UCB-TH} achieve the best results among all tasks, and the former is slightly better than the latter in most tasks. It indicates that \textit{UCB-Beta} and \textit{UCB-TH} might be equivalent in our setting. \textit{UCB-TN} is worse than \textit{UCB-Beta} and \textit{UCB-TH}, indicating that using a Gaussian distribution to fit the cluster distribution might be inaccurate. Although \textit{UCB1} performs better than \textit{Random-Draw}, it is far worse than \textit{UCB-Beta}, \textit{UCB-TH}, and \textit{UCB-TN}, showing that only estimating the mean of the distribution of each cluster could not solve the inter-cluster data selection problem. 

%% file: latex/conclusion.tex
\section{Conclusion}
In this paper, we aim to reduce the computational consumption used in gradient-based SFT data selection for LLMs. Our proposed framework first performs clustering over the training data pool based on the intuition that training data samples with similar gradients would have similar influences on target loss optimization. Then, we frame the inter-cluster data selection as a computing budget allocation problem which is similar to the multi-armed bandit problem, and modify the UCB algorithm to solve it. Combined with the state-of-the-art gradient-based data selection methods, experimental results show that our proposed framework can match the original methods while greatly reducing the computing consumption. 

\section*{Limitations}
While our proposed framework has been proven to be efficient in saving computing resources, it is also essential to consider its limitations that may be improved in the future. As stated by \citet{wang5206107dynamic}, the gradient-based data selection methods only consider the influence of single data samples, neglecting the mutual influences within the selected data subsets. With the clusters generated, the inter-cluster data selection could consider groups of data samples as the arm-drawing rewards, which is the next step of our work. Moreover, we use K-means, the simplest clustering algorithm, in this paper. Better clustering might also improve the performance of our proposed framework.

%% file: latex/appendix.tex
\section{Training and evaluation}

\subsection{Training datasets}

We use the same training datasets as \cite{wang5206107dynamic} did. The number of instances in each dataset are shown in Table~\ref{tab:dataset}. The number of averaged completion tokens in our training data pool is 189.1. Setting the selection ratio to 5\%, the selected data subset contains 20387 instances.

\subsection{Training details}
Following \citet{xia2024less} and \cite{wang5206107dynamic}, we adopt the parameter-efficient fine-tuning method LoRA~\cite{hu2022lora} in all our experiments. The rank of LoRA module is 128, the value of $\alpha$ is 512, and the learnable LoRA matrics are applied to all attention matrices. Under this configuration, there are 134,217,728 trainable parameters in LLaMA-2-7B accounting for 1.95\% of the original parameters, and 58,982,400 trainable parameters in Qwen2.5-3B accounting for 1.88\% of the original parameters. All of our experiments are conducted using 8 Tesla V100 GPUs.

\subsection{Evaluation details}
We use the evaluation code toolkit provided by Open-Instruct~\cite{wang2023far}. On MMLU, the evaluation metric is the exact match of the first token in models' completion and the ground truth answer, we perform evaluation in a 5-shot setting and average over the 57 subtasks; On TydiQA, gold passage and 1-shot are adopted, and the performance is evaluated as the F1 score of the models' completions and the ground truth answers and averaged over nine languages; On GSM8k, 8-shot is adopted, and the final number in models' completion is extracted as the final answer to exactly match with the ground truth answer; On HumanEval, we use pass@1 as the evaluation metric and sample 20 completions for each instruction with temperature 0.1.

\begin{table}[t]
    \centering
    \begin{tabular}{lc}
        Dataset & \# Instance \\
        \hline
        Flan v2 & 99245  \\
        CoT & 95557  \\
        Dolly & 14865 \\
        Open Asisstant v1 & 54626  \\
        GPT4-Alpaca & 52002  \\
        ShareGPT & 63951  \\
        GSM8k train & 7473  \\
        Code-Alpaca & 20021  \\
        \hline
        Total & 407740
    \end{tabular}
    \caption{The number of instances in each dataset used in our experiments.}
    \label{tab:dataset}
\end{table}

\begin{figure}
    \centering
    \includegraphics[width=1\linewidth]{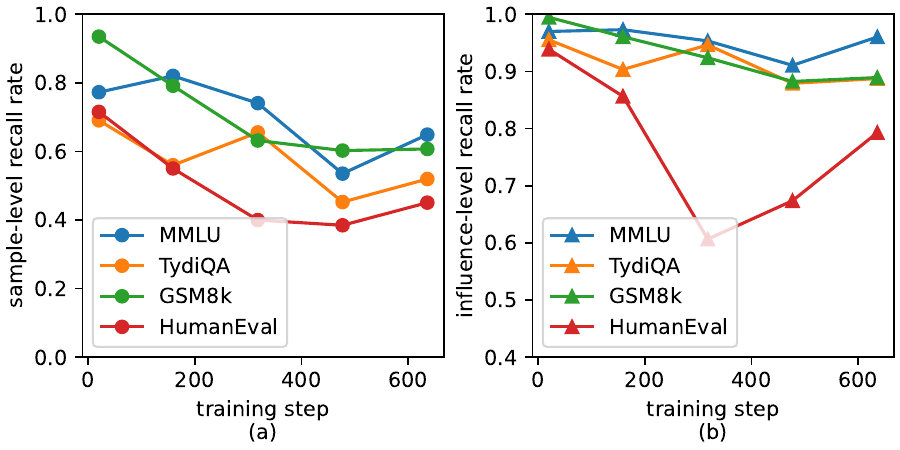}
    \caption{The change of sample-level and influence-level recall rates during the training process.}
    \label{fig:intra-cluster}
\end{figure}

\section{Declination of the effectiveness of clusters during training}
\label{app:cluster}

In our proposed framework, we only perform clustering according to the cosine similarities of the gradients of training data samples at the beginning of training. Since the gradients of training data samples would change with the update of model weights, whether the clustering is still effective during the training process would be an essential problem. Thus, we conduct experiments to study the changing trend of the effectiveness of clusters during training.

We again use the sample-level recall rate $R_{s}$ in Equation~\ref{eq:recall} and influence-level recall rate $R_{inf}$ in Equation~\ref{eq:score} to evaluate the effectiveness of clusters. We compute $R_{s}$ and $R_{inf}$ with respect to the checkpoints saved during training with 5\% randomly selected data samples. The number of clusters $k$ = 150, the cold start ratio $p_{cs}\%$ = 5\%, as in our main experiments. The results are illustrated in Figure~\ref{fig:intra-cluster}. 

On most benchmarks, $R_{s}$ and $R_{inf}$ show declining trends when the training step increases. This indicates that updating clustering after certain training steps could lead to better results. However, the updating of clusters also introduces extra computational consumption. Moreover, $R_{inf}$ still remains high in the later stage of training. Hence, we choose not to update the clustering in our experiments. Still, we obtain comparable results with methods using the full budget according to Table~\ref{tab:main}.

\section{Exploration vs. exploitation of inter-cluster data selection}
\label{app:dist}

\begin{figure*}
    \centering
    \begin{subfigure}{\linewidth}
        \centering
        \includegraphics[width=\linewidth]{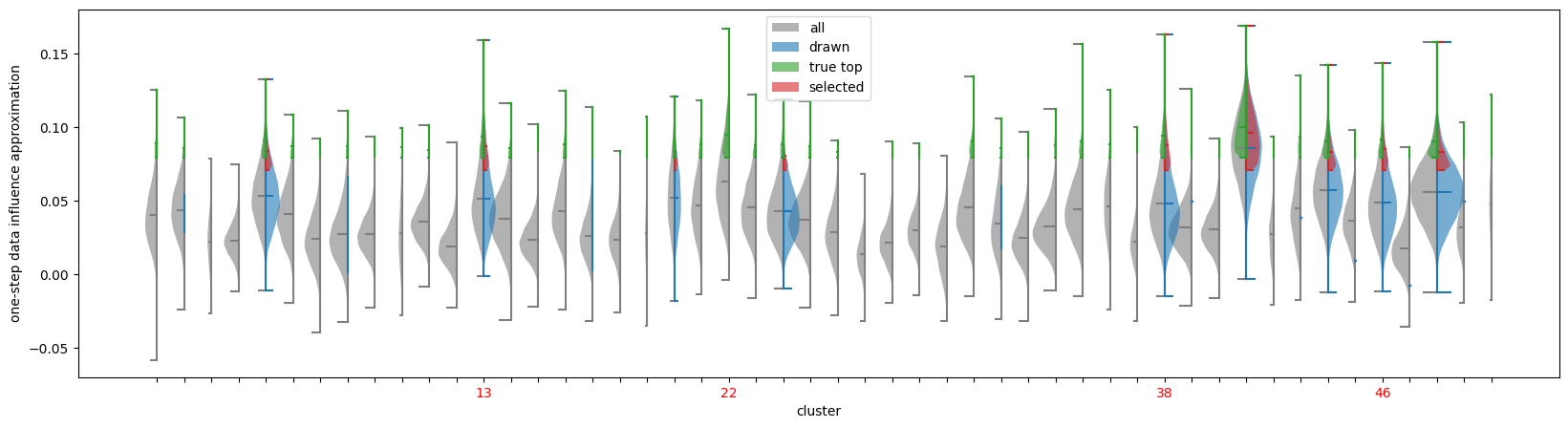}
        \caption{cold start ratio = 0\%}
        \label{subfig:csp0}
    \end{subfigure}
    \begin{subfigure}{\linewidth}
        \centering
        \includegraphics[width=\linewidth]{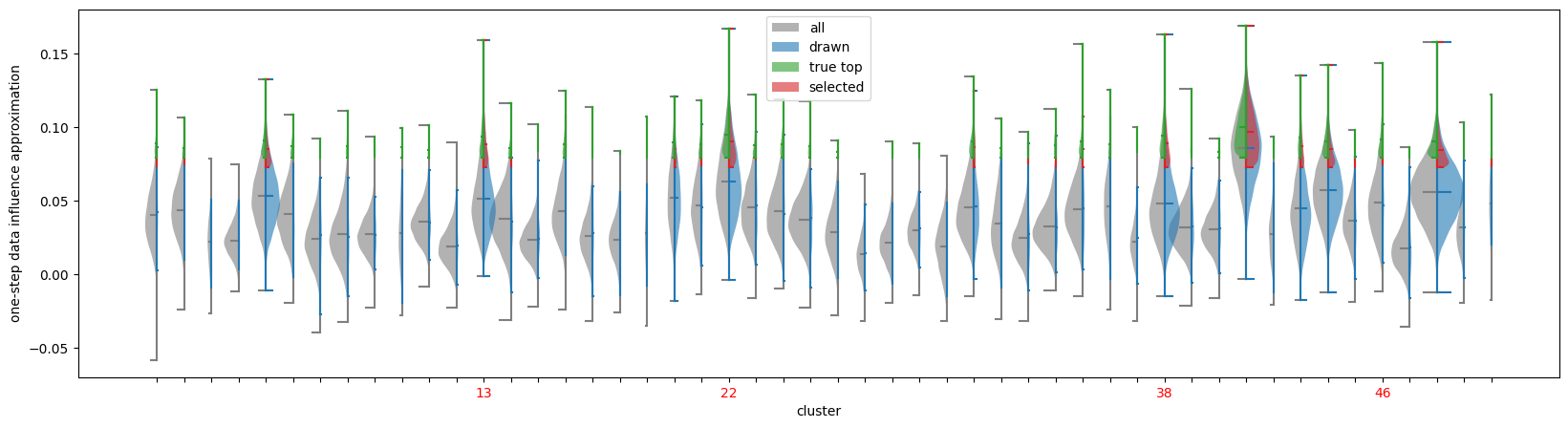}
        \caption{cold start ratio = 5\%}
        \label{subfig:csp5}
    \end{subfigure}
    \begin{subfigure}{\linewidth}
        \centering
        \includegraphics[width=\linewidth]{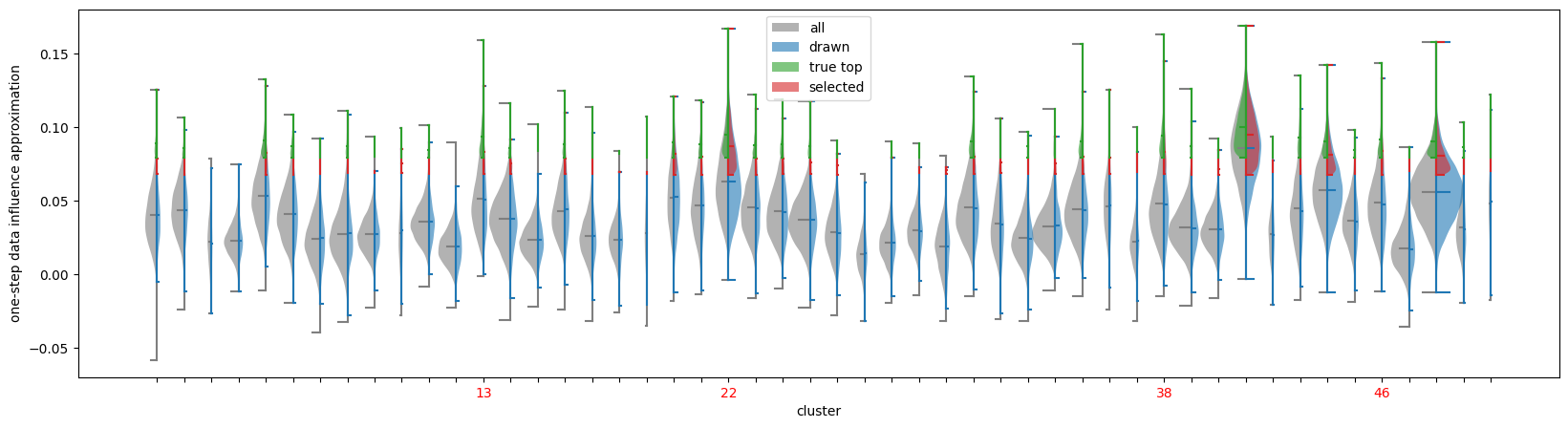}
        \caption{cold start ratio = 50\%}
        \label{subfig:csp50}
    \end{subfigure}
    \caption{Data distributions among clusters with different cold start ratios on MMLU benchmark. Each violin graph represents one cluster. For each cluster, the gray half is the total data samples contained in this cluster; the blue half is the data samples drawn from this cluster in our UCB algorithm; the green half is the true top portion of data samples with the highest influences contained in this cluster; the red half is the selected data samples from this cluster using ClusterUCB. The width of each half represents the number of data samples in this half.}
    \label{fig:dist}
\end{figure*}

As discussed in Section~\ref{subsubsec:csp}, the cold start ratio $p_{cs\%}$ controls the trade-off between exploration and exploitation in our UCB algorithm. To further observe the effect of this trade-off, we plot the distribution of the total data samples contained, the data samples drawn in our UCB algorithm, the true top portion of data samples with the highest data influence approximations, and the selected data samples using our modified UCB algorithm within each cluster. For simplicity, we plot with the number of clusters $k$ = 50 on the MMLU benchmark. To compare the effect of different degrees of exploration and exploitation, we plot with the cold start ratio $p_{cs}\%$ = 0\%, 5\%, and 50\%, as shown in Figure~\ref{subfig:csp0},~\ref{subfig:csp5} and~\ref{subfig:csp50}, respectively.

When $p_{cs}\%$ = 0\%, the modified UCB algorithm does not adopt the cold start strategy and tends to assign most of the computing budget to exploitation. Accordingly, the distribution of the drawn data samples in Figure~\ref{subfig:csp0} is more concentrated on a few clusters, which does not cover many clusters with high-influence data samples. Assigning a small budget to random exploration with $p_{cs}\%$ = 5\%, the evaluation of each cluster is more accurate in our UCB algorithm, leading to larger probability to find clusters with more high-influence data samples, e.g., cluster No. 22 and 46 in Figure~\ref{subfig:csp5}. Continue increasing $p_{cs}\%$ to 50\%, more budget is spend on exploration, resulting in insufficient budget for exploitation, the modified UCB algorithm is more likely to miss high-influence data samples even though it can hit the corresponding clusters, e.g., cluster No. 13 and 38 in Figure~\ref{subfig:csp50}.

Thus, the distribution of inter-cluster data selection is consistent with our experimental results in Section~\ref{subsubsec:csp}, that a cold start with random exploration is necessary in the inter-cluster data selection, but spending too much budget on exploration could be harmful and lead to worse performance.

